\documentclass[10pt,twocolumn,letterpaper]{article}

\usepackage[pagenumbers]{cvpr} 

\usepackage{graphicx}
\usepackage{babel,blindtext}
\usepackage{mathtools}
\usepackage{amssymb}
\usepackage{caption}
\usepackage{subcaption}
\DeclareMathOperator*{\argmin}{arg\,min} 
\usepackage{graphicx}
\usepackage[toc,page]{appendix}

\graphicspath{ {./images/} }

\usepackage[pagebackref,breaklinks,colorlinks]{hyperref}

\usepackage[capitalize]{cleveref}
\crefname{section}{Sec.}{Secs.}
\Crefname{section}{Section}{Sections}
\Crefname{table}{Table}{Tables}
\crefname{table}{Tab.}{Tabs.}

\def\cvprPaperID{*****} 
\def\confName{CVPR}
\def\confYear{2022}
\usepackage{cite} 
\usepackage{url} 
\usepackage[utf8]{inputenc} 
\usepackage{booktabs} 

\begin{document}
\title{The Heat is On: Thermal Facial Landmark Tracking}


\author{James Baker \\
Department of Computer Science\\
University of Houston\\
}
\date{25/4/15}

\maketitle

\begin{abstract}
Facial landmark tracking for thermal images requires tracking certain important regions of subjects' faces, using images from thermal images, which omit lighting and shading, but show the temperatures of their subjects. The fluctuations of heat in particular places reflect physiological changes like bloodflow and perspiration, which can be used to remotely gauge things like anxiety and excitement. Past work in this domain has been limited to only a very limited set of architectures and techniques. This work goes further by trying a comprehensive suit of various models with different components, such as residual connections, channel and feature-wise attention, as well as the practice of ensembling components of the network to work in parallel. The best model integrated convolutional and residual layers followed by a channel-wise self-attention layer, requiring less than 100K parameters.

\end{abstract}

\section{Introduction}
Detecting physiological changes in human faces, like bloodflow or perspiration, can measure stress \cite{shastriImaging2009, shastriperinasal2012}, empathy \cite{EBISCHmother2012}, and even deceit \cite{tsiamyrtzisimaging2007}. Thermal images are ideal for conveying physiological changes. They are non-invasive, unlike sensors, and are sensitive to the heat changes caused by physiological changes, unlike normal images. However, using thermal images requires consistently measuring the same region of interest on a human subject, which is complicated by the fact that humans move, tilt and rotate their bodies. It is imperative to use a robust, reliable framework for landmark tracking for these tasks.

Previous work on thermal image facial landmark tracking did not implement techniques such as non-locality, residual layers and ensembling or ''wisdom-of-crowds''. This paper rectifies that by using hyperparameter optimization to construct models with residual and attentional elements. Multiple, parallel components were added to the top models, in order to test if ensembling would improve performance.

\subsection{Contributions of This Work}
After experimentation, the key insight from this work was that Convolutional Neural Networks that used attention and residual components were not only accurate, but also very lightweight in their number of parameters, which will assist in their use in smaller, mobile devices. In fact, increasing their complexity by adding parallel components made the best models overfit. 

\section{Related Work}
In affective computing, researchers need to measure physiology of their subjects. Unfortunately, this often requires attaching sensors to people, which is burdensome to the subjects and does not scale economically. Using cameras solves this, but non-thermal images do not provide any information on temperature. Dusty, foggy or dark environments may make rob normal images of any information, and normal cameras will barely pick up anything, whereas thermal imaging is still robust under bad conditions.

The specific task we are interested in is landmark tracking. Given a face, we want to reliably find specific objects or regions (landmarks), like the eyes, nose or lips. There have been many datasets made for this task for normal images \cite{Shen2015TheFF,300W}. What makes thermal facial images unique is skin temperature changes in response to things like bloodflow or breathing, which means that regions of interest may be prone to suddenly blend in or stand out or change color over time. Thermal images are also difficult as they tend to lack the textural information in normal images \cite{PosterExamination}. While they offer more information in some ways, they are challenging in others, thus using thermal images is a ''higher risk, higher reward'' version of working with normal images.

\subsection{Statistical Methods}
An adjacent task to landmark detection is region-of-interest tracking. Using videos of faces, \cite{Zhou2013SpatiotemporalSA} used a particle filter tracker in order to follow regions of interest. Similarly, \cite{Mostafa2013} used a particle filter and an object detector that follows the position of the eyes to build features for classification of regions using a random fern algorithm. Other landmark detection has been in the domain of normal images. 
Many methods use a regression framework, which can be formulated like so.
Given a set of p coordinates of landmarks at time t \(S_t\):
\begin{equation}
  S_t=\{(x_0,y_0), (x_1,y_1)...(x_p,y_p)
  \label{eq:landmarks}
\end{equation}
Given regressor \(r_t\) and image \(I\), we can model:
\begin{equation}
S_{t+1}=S_t+r_t(I,S_t)
  \label{eq:reg}
\end{equation}
Given predicted \(\hat{S_t}\) and actual \(S_t^*\), we choose \(r_t\) as the solution to the optimization problem
\begin{equation}
\argmin_{r_t}=\sum^N || S_t^*-\hat{S_t}-r_t||_2
  \label{eq:optim}
\end{equation}
In \cite{Sanchez_Lozano_2018}, the authors used a continuous regression method, where the input space was approximated using Taylor Expansions, which proved to be robust to translation, tilts and other shifts, as well as computationally efficient to solve for. \cite{Hao2017SalientpointsguidedFA} also used a cascaded regression to find landmarks. Then, using the location of landmarks, they found similar faces in the dataset and used those to fit a regression for each face. \cite{martinezlocalevidence} used an approach where at each step, the sampling region for each landmark was updated, and a Markov Random Field ensured that the new sampling regions were consistently in a face shape. \cite{Yang2013SievingRF} used a random regression forest, followed by a cascade of sieves to filter out votes that are too inconsistent or distant from the hypothesis for the predicted landmark. 
\subsection{Deep Learning Methods}
\cite{Keong2020MultispectralFL} used the U-Net, introduced by \cite{RonnebergerFB15}, which is a deep convolutional neural network (CNN) that first downsamples and then upsamples the input images, with a fully connected head. The CNN was trained, and then the fully connected layer. They used one U-Net to find a square enclosure of the face, and passed the enclosed region to another U-Net that predicted the coordinates of the landmarks. \cite{ChuDeepMultiTask} expanded on the sequential U-Net model by replacing the final fully connected layer with two parallel layers, one for landmark detection and one for emotion classification. \cite{PosterExamination} compared three different CNN models for landmark detection: a Multitask Cascaded Convolutional Network \cite{zhengMTCCN}, a Deep Alignment Network \cite{KowalskiDAN}, and a Multiclass Patch Based Classifier, a CNN which for every 60 x 60 patch of the input image classifies it as one of six regions.

\section{Proposed Approach}

\subsection{Singular Models} \label{singular}
All models consisted of 4 basic components. The first was a ``root'', which was the same in all models, that consisted of 2 convolutional layers for downsampling the input while also increasing the number of feature channels, translating the images \(\mathbb{R}^{480 \times 640 \times 3} \rightarrow \mathbb{R}^{120 \times 160 \times 64}\). This was also done because models that were initially convolutional but then implemented attentional layers later on had shown to perform better \cite{standaloneramachandran} and are used in other successful experiments \cite{xu2016show}. Following the root was a ''stem'', then  an optional ''branch''. Following the stem, or branch if the model had one, was a ''head'', which flattened the outputs of the preceding layer,applied a fully connected layer with 2N nodes and dropout, where N is the amount of points to be predicted, and then reshaped the outputs from \(\mathbb{R}^{2N} \rightarrow \mathbb{R}^{2 \times N}\).

A stem was made by repeating one of five different layers. More detail describing :
\begin{itemize}
    \item Convolutional layers (referrred to as Conv Stem)
    \item ResNeXt layers (referred to as ResNeXt stem)
    \item Alternating Convolutional layers and Bahdanau feature-wise attention layers (referred to as Alternating Conv-Bahdanau Stem)
    \item Alternating Convolutional layers and Luong feature-wise attention layers (referred to as Alternating Conv-Luong Stem)
    \item Alternating Convolutional layers and ResNeXt layers (referred to as Alternating Conv-ResNeXt stem)
\end{itemize}
A branch was made by repeating one of four different layers
\begin{itemize}
    \item Nothing; the stem was directly connected to the head (referred to as No Branch)
    \item Luong feature-wise attention layers (referred to as Luong branch)
    \item Bahdanau feature-wise attention layers (referred to as Bahdanau branch)
    \item A single Patch Encoder layer (not repeated), and then Transformer spatial-wise attention layers (referred to as Vision Transformer branch)
\end{itemize}
\begin{figure*}[h]
    \centering
    \includegraphics[scale=.75]{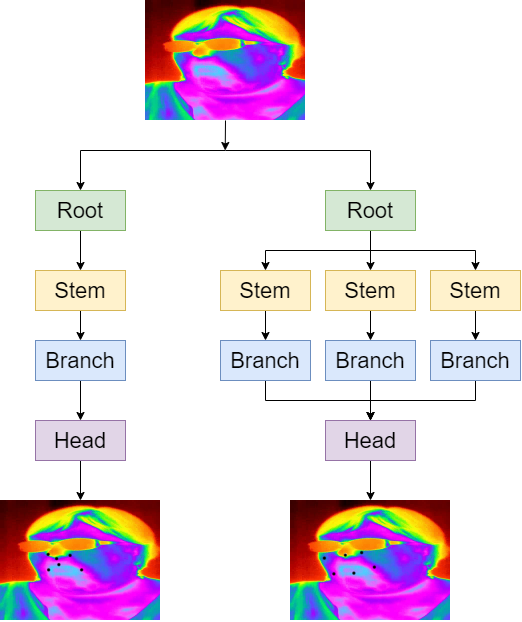}
    \caption{Singular Model (left), and Ensemble Model (right), in testing mode, with Image 8282 from subject 64}
    \label{fig:singular}
\end{figure*}

\subsubsection{Optimization}

Due to the massive search space of hyperparameters (depth, kernel size, etc.) for these models, we used the Optuna library \cite{optunaakiba} to efficiently and automatically run trials to find the optimal hyperparameters. For searching for which hyperparameters to test for each trial, we used the Tree-Structured Parzen Estimator (TPE), given it had performed well experimentally \cite{Olof2018ACS}. The TPE algorithm is a greedy algorithm that samples hyperparameters for new trials from a distribution estimated using the hyperparameters from past previous trials \cite{bergstraoptimization2011}. For pruning unpromising trials, we used the Asynchronous Successive Halving (ASHA), which also performed well in experiments \cite{li2020massively}. ASHA works by asynchronously promoting the best trials based on some performance threshold, and then canceling the rest.

For each stem, for each branch, we ran 40 trials that ran for 20 epochs each, and chose the model configurations that returned the lowest loss to be the optimal, leading to 20 different models.
The loss function that was to be minimized was wing loss \cite{fenfwingloss}, which penalizes small and medium weight errors, and had been shown to perform better for facial landmark localisation tasks. The size (in quantity of parameters) and wing loss of the final best models is given in \cref{optunasingulartable}. The singular models have been indexed with arbitrary letters of the alphabet, and in later tables, models will be referred to using their alphabetical letter. 

\begin{table*}[h] 
\centering 
\begin{tabular}{ l | c | c } 
\toprule
Model & Params & Wing Loss \\
\midrule
 C-1: Conv Stem + No Branch & 383100 & 2.0686  \\
 C-2: Conv Stem + Luong Branch & 139404  & 4.1657\\
 C-3: Conv Stem + Bahdanau Branch & 533948 & 1.4898  \\
 C-4: Conv Stem + Vision Transformer Branch & 215868  & 1.7841\\
 \hline
 R-1: ResNeXt Stem + No Branch & 287792 & 3.3708\\
 R-2: ResNeXt Stem + Luong Branch & 294844  & 3.5428\\
 R-3: ResNext Stem + Bahdanau Branch &  2275528 & 1.6349\\
 R-4: ResNeXt Stem + Vision Transformer Branch & 159420  & 2.4904\\
 \hline
 L-1: Alternating Conv-Luong Stem + No Branch & 618844  & 1.7184\\
 L-2: Alternating Conv-Luong + Luong Branch & 303836 & 4.6675 \\
 L-3: Alternating Conv-Luong + Bahdanau Branch & 283772  & 1.5763\\
 L-4: Alternating Conv-Luong + Vision Transformer Branch & 911292  & 2.306\\
 \hline
 B-1: Alternating Conv-Bahdanau Stem + No Branch & 592748  & 2.7187\\
 B-2: Alternating Conv-Bahdanau + Luong Branch & 262028  & 5.092\\
 B-3: Alternating Conv-Bahdanau + Bahdanau Branch & 599276 & 2.5331 \\
 B-4: Alternating Conv-Bahdanau + Vision Transformer Branch & 747948 & 2.5257 \\
 \hline
 A-1: Alternating Conv-ResNeXt Stem + No Branch & 762052 & 1.3416 \\
 A-2: Alternating Conv-ResNeXt + Luong Branch &  91280 & 2.5564\\
 A-3: Alternating Conv-ResNeXt + Bahdanau Branch & 541380 & 1.1292 \\
 A-4: Alternating Conv-ResNeXt + Vision Transformer Branch & 2070140 & 1.3802 \\
 \hline
 \end{tabular}
\smallskip 
\caption{Optimal Models from Optuna}
\label{optunasingulartable}
\end{table*}

\subsection{Ensemble Models}
The success of ensemble learning, using multiple models together for the same task, has proven successful in fields such as robotics \cite{Zhang2020GraspFS}, finance \cite{Yang2020DeepRL} and medicine \cite{ALI2020208}, as ensembling reduces variance \cite{Breiman96bias} and evades local optima \cite{ensemblereview}. Specifically, this paper used stacked generalizations \cite{WOLPERT1992241}, where a final algorithm is trained to use the outputs of the component models ''stacked'' together. In this case, this was implemented by concatenating the flattened outputs of all the component models and applying a fully-connected head. Each of the component models also shared a convolutional root, as using the same root would add parameters, and each root would likely be supplying redundant information.

\section{Experiments} 
\subsection{Implementation Details}
\begin{figure*}[h]
\includegraphics[scale=.7]{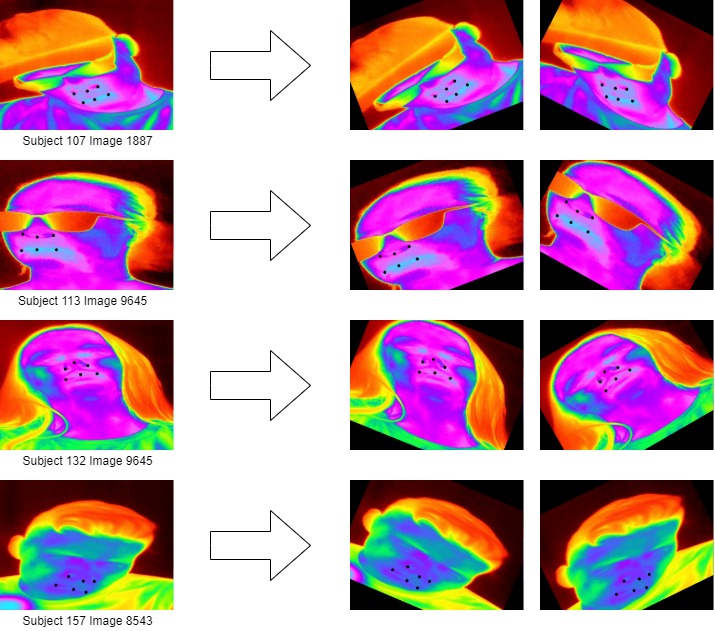}
\caption{Original and Augmented Images, Annotated with locations of landmark points}
\label{fig:concat}
\end{figure*}

The data consisted of roughly 2000 thermal images collected from videos with 73 different subjects, each of which had been annotated with the locations of 16 landmarks. Ten of the videos were collected as part of an IRB approved study. The subjects were recruited through email and personal communications—the subjects work/study at a public research university. They were filmed in an office setting doing work like grant proposals and writing papers. They were filmed visually using a Tau 640 long-wave infrared (LWIR) camera (FLIR Systems, Wilsonville, OR. The camera features 50◦ mK thermal resolution, 640 × 480 pixels spatial resolution, and an auto-focus mechanism. The camera was located under the participant’s desktop screen, attached to a Bescor MP- 101 Motorized Pan; Tilt Head (Bescor, Farmingdale, NY) to facilitate face tracking. Thermal facial data were collected at a frame rate of approximately 30 fps using a 35 mm lens. The other 63 subjects were knowledge workers and were recorded using a Tau 640 long-wave infrared (LWIR) camera (FLIR Systems, Wilsonville, OR), featuring a small size (44 × 44 × 30 mm) and adequate thermal (50 mK) and spatial resolution (640×512 pixels), though the images were cropped to 640 × 480 pixels for this experiment. A LWIR 35mm lens f/1.2, controlled by a custom auto-focus mechanism was fitted on the camera. The thermal camera is located under the participant’s computer screen, attached to Bescor MP-101 Motorized Pan and Tilt Head (Bescor, Farmingdale, NY) to facilitate face tracking. The 63 images were previously published \cite{Zaman2019}. As the upper lip region is the most useful for stress studies \cite{shastriImaging2009, shastriperinasal2012}, we only trained the Model to find the 6 points surrounding that region. In order to make the data more robust, each image and the relevant landmarks were rotated by a random angle between 20 and 30 degrees around the center of the drawing, to the left and the right, as shown in \cref{fig:concat}. 

\subsection{Results}
\subsubsection{Singular Results}
A total of 20 different models singular models were tested. The exact hyperparameters were found using the methods discussed in \cref{singular}. Each of these models was trained for 100 epochs with a batch size of 32. \cref{batch32} reports the Accuracy, Wing Loss, Mean Absolute Error (MAE) and Mean Squared Error (MSE) of each model when tested.

As \cref{batch32} shows, the best models were those that alternated ResNeXt and Convolutional layers.

\begin{table}[h]
\centering 
\begin{tabular}{l | c c c c } 
\toprule 
& \multicolumn{4}{c}{metrics} \\ 
\cmidrule(l){2-5} 
 Model & Accuracy  & Wing Loss  & MAE &  MSE \\ 
\midrule 
Model C-1  &  0.9389 & 1.7195 & 0.0292 & 0.0018  \\
Model C-2 & 0.9182 & 2.2061 & 0.0376 & 0.0026 \\
Model C-3 & 0.9482 & 1.4323 & 0.0244 & 0.0012 \\
Model C-4 & 0.9517 & 1.3282 & 0.0226 & 0.0011 \\ 
\hline
Model R-1 & 0.6848 & 21.9716 & 0.4085 & 0.1873 \\
Model R-2 & 0.5723 & 21.9296 & 0.4054 & 0.1846 \\
Model R-3 & 0.6188 & 22.1513 & 0.4096 & 0.1885 \\
Model R-4 & 0.679 & 22.1508 & 0.4103 & 0.19 \\
\hline
Model L-1  & 0.9541 & 1.2664 & 0.0209 & 0.001 \\
Model L-2 & 0.8572 & 3.7362 & 0.0649 & 0.0075 \\
Model L-3 & 0.9532 & 1.277 & 0.0214 & 0.001 \\
Model L-4 & 0.9446 & 1.5254 & 0.0259 & 0.0014 \\
\hline
Model B-1 & 0.9193 & 1.9792 & 0.0339 & 0.0024 \\
Model B-2 & 0.9078 & 2.4484 & 0.0418 & 0.0033 \\
Model B-3 & 0.9146 & 2.2086 & 0.0371 & 0.0028 \\
Model B-4 & 0.9262 & 1.9015 & 0.0325 & 0.0022 \\
\hline
Model A-1 & 0.966 & 0.9175 & 0.0153 & 0.0005 \\
Model A-2 & 0.96 & 1.0997 & 0.0188 & 0.0007 \\
Model A-3 & 0.9715 & 0.7628 & 0.0128 & 0.0003 \\
Model A-4 &  0.9538 & 1.0896 & 0.0186 & 0.0007 \\
\midrule 
\end{tabular}
\smallskip 
\caption{Singular Model Results} 
\label{batch32}
\end{table}

Model A-3 stands out as the best one. It achieves the lowest wing loss of 0.7628, while using only 91,280 parameters, while most state of the art models use millions.

\subsubsection{Ensemble Results}
Given that the Alternating Conv-ResNeXt models (Models Q,R,S,T) performed best, we then tried to improve them by using an ensemble of 3 parallel stem-branch components connecting the root to the head. Results are shown in \cref{tab:ensembleresults}. Performance consistently decreased with the use of ensembling.

\begin{table}[h]
\centering 
\begin{tabular}{l | c c c c } 
\toprule 
& \multicolumn{4}{c}{metrics} \\ 
\cmidrule(l){2-5} 
 Model & Accuracy  & Wing Loss  & MAE &  MSE \\ 
\midrule 
    Model A-1 & 0.9643 & 0.9921 & 0.0167 & 0.0008 \\
    Model A-2 & 0.9592 & 1.153 & 0.0197 & 0.0009 \\
    Model A-3 & 0.6812 & 21.9525 & 0.407 & 0.1864 \\
    Model A-4 & 0.5726 & 22.2677 & 0.4137 & 0.1914\\
        \end{tabular}
    \caption{Ensemble Model Results}
    \label{tab:ensembleresults}
\end{table}

\subsection{Ablation Study}
\subsubsection{Data Without Rotations}
In order to test if rotating the data was actually useful, we then repeated the experiments with the alternating residual and convolutional singular models. Given that the data set was one third of the size without the rotated images, we trained them for 300 epochs instead of the usual 100. Results are shown in \cref{tab:ablation}. There were significant decreases in performance, as predicted.
\begin{table}[h]
    \centering
\begin{tabular}{l | c c c c } 
\toprule 
& \multicolumn{4}{c}{metrics} \\ 
\cmidrule(l){2-5} 
 Model & Accuracy  & Wing Loss  & MAE &  MSE \\ 
\midrule 
Model A-1 & 0.9474 & 1.3288 & 0.0225 & 0.0012' \\
Model A-2 & 0.9389 & 1.4316 & 0.0243 & 0.0014' \\
Model A-3 & 0.9599 & 1.0501 & 0.0176 & 0.0008 \\
Model A-4 & 0.9523 & 1.5673 & 0.0263 & 0.0016 \\
    \end{tabular}
    \caption{Models Trained Without Rotations}
    \label{tab:no_rotations}
\end{table}
\subsubsection{Removing Layers}
To test if the models were too complex, we removed the last convolutional and ResNeXt layer. As \cref{tab:ablation} shows, wing loss increased in all cases but the Model A-4 (alternating convolutional and ResNeXt stem and a Vision Transformer branch), in which case performance stayed nearly the same.
\begin{table}[h]
    \centering
\begin{tabular}{l | c c c c } 
\toprule 
& \multicolumn{4}{c}{metrics} \\ 
\cmidrule(l){2-5} 
 Model & Accuracy  & Wing Loss  & MAE &  MSE \\ 
\midrule 

Model A-1 & 0.7341 & 18.0931 & 0.3301 & 0.1339 \\
Model A-2 & 0.9553 & 1.1807 & 0.0192 & 0.0008 \\
Model A-3 & 0.9654 & 0.9057 & 0.0152 & 0.0005 \\
Model A-4 & 0.9632 & 1.0893 & 0.0185 & 0.0007 \\
    \end{tabular}
    \caption{Models with Layers Removed}
    \label{tab:ablation}
\end{table}
 
\subsection{Activation Maximization}
Activation Maximization consists of searching for an input that produces the maximum possible value for some particular output of the model \cite{MONTAVON20181}. This can mean maximizing the output of a particular unit or even a whole layer in a network. Activation Maximization is useful for visualizing what features hidden layers are responsive to.
\subsubsection{Gradient Ascent}

In this case, we used gradient ascent \cite{mahendran2014understanding} on a random noise image to maximize the output of particular channels of output layers. The outputs were extremely abstract and surreal. In Appendix A, \cref{fig:resnext_layer_4}  and  \cref{fig:resnext_layer_5} show samples of images that were made to maximize different output channels in the  second to last layer, and last layer, respectively, of the alternating convolutional and residual singular model.

\section{Conclusions and Recommendations}
In this paper, we demonstrated that models that had alternating convolutional and residual components in early layers, with attentional components in later models, were very effective in facial landmark tracking for thermal images. Specifically, the best model (Model A-3) used alternating residual and convolutional layers, and then a luong channel-wise attention layer, while using less than 100,000 parameters. This is not surprising given that one of the selling points of the ResNeXt layer \cite{Xie2017AggregatedRT} was its high performance relative to its small number of parameters.

The failure of ensembling is likely due to the ability of models that are too complicated learning to overfit the training data, resulting in high generalization error \cite{belkin2019reconciling}. Even standard forms of regularization employed here like dropout and data augmentation may not sufficiently correct this \cite{zhang2017understanding}.

This particular task of locating this particular region of the face is useful for monitoring stress, which can affect whether we operate machinery \cite{panagapoulosforecasting2020}, drive automobiles \cite{pavlidisbiofeedback2021,gomezcausality2018} safely, and even perform surgery \cite{Pavlidis2019AbsenceOS}. Ideally, this work can be used for further research on how humans perform under stress and monitoring stress as we perform tasks to gauge when we are prone to make potentially dangerous mistakes. Sleep studies also require unobtrusive facial measurements, and this work may also be useful for those as well.

\subsection{Limitations}
The experiments in this paper were done on only one dataset. Our model may not robust to facial attributes not present in this dataset (some people only have one eye, but none of the subjects did). The images were also very high-quality; the assumption is that this would be used for scenarios where humans were very close to the cameras. For example, when operating an automobile, or in a lab, it would be feasible to keep a heat camera a few feet from the humans face. The subjects were also not filmed doing physically strenuous tasks like exercising or manual labor. The increased blood flow and perspiration may create very different heat patterns in subjects faces.

{\small
\bibliographystyle{ieee_fullname}
\bibliography{sources}
}

\appendix
\section{Components} \label{sec:appendixA} 
\subsection{Convolution}
Generally, a convolution of two functions is their product over a range \cite{wolframconvolution}.  In the continuous case, this can be expressed as an integral, like so:
\begin{equation}
[f * g] (t)=\int f(t) g(t-\tau) d \tau
  \label{eq:conv}
\end{equation}
In the discrete case, this can be expressed as a sum, like so:
\begin{equation}
[f * g] (t)=\sum_{\tau} f(t) g(t-\tau)
  \label{eq:convdiscrete}
\end{equation}
The sums can be across multiple axes
\begin{equation}
[f * g] (t,s,r)=\sum_{\tau} \sum_{\sigma} \sum_{\rho} f(t,s,r) g(t-\tau,s-\sigma,r-\rho)
  \label{eq:convmultiple}
\end{equation}
A 2D Convolution of an image \(I\), that can be represented as an \(H \times W \times C\) tensor (usually for black and white images, C=1, for color images C=3), and a kernel \(K\), that can be represented as an \(H_k \times W_k \times C\) tensor, (where typically \(H_k <<< H\) and \(W_k <<< W\)) is usually of the form:
\begin{equation}
\mathbf{Conv2D}(x,y)=\sum_c^C \sum_h^{H_k} \sum_w^{W_k} I_{x,y,c} K_{x-h,y-w,c}
  \label{eq:conv2d}
\end{equation}
Often, convolutions do not sample every possible \(x,y\) value. The 'stride' denotes the amount of pixels between each pixel to apply the convolution function to. For example, with a \(stride=(2,2\), we would only perform:
\(\mathbf{Conv2D}(x,y),\mathbf{Conv2D}(x+2,y),\mathbf{Conv2D}(x,y+2),\mathbf{Conv2D}(x+2,y+2)\), skipping intermediate values of \(x,y\). Thus, each 2D Convolution produces a new \(H'' \times W''\) output map. If there are \(C''\) different kernels (all with the same shape), then we can concatenate all of the output maps to create a \(H'' \times W'' \times C''\) tensor. The weights of each kernel are tunable parameters that are optimized during training \cite{Goodfellow-et-al-2016}. 
\subsection{Attention}
When humans process input from a sequence or an image, we contextualize each part of the input using other parts of the input. However, we do not pay equal attention to every other part of the input \cite{MnihHGK14}. For a textual example, in the sentence ''Alan said he was hungry'', the word ''alan'' is more useful in determining the meaning of ''he'' than the word ''hungry''. For a visual example, in order to guess the location of the right eye on someones face, the most important information would be the location of the left eye, not the length of their beard. Attention, also known as Non-Locality was first applied to text sequences, for neural machine translation \cite{LuongPM15,bahdanau2016neural,GravesWD14}. Later, networks with attention were used for image generation \cite{GregorDGW15draw}, video captioning \cite{XuBKCCSZB15showattendtell} and object recognition \cite{ba2015multiple}.
Attention \cite{yin2020disentangled} is formulated as
\begin{equation}
Attention(Q,K,V)=\mathrm{softmax}(\mathrm{score}(Q,K))V
  \label{eq:attention}
\end{equation}
\(Q,K,V\) are query, key and value matrices. When \(Q=K=V\), this is called self-attention. Softmax \cite{Nwankpa} is defined as:
\begin{equation}
\mathrm{softmax}{ (x_i)}=\frac{\exp{x_i}}{\sum_j \exp{x_j}}
  \label{eq:softmax}
\end{equation}
In \cite{LuongPM15}, they represent the hidden states of each sequence of the input, derived from the output of the LSTM layers that precede the attentional layers in the model. In other contexts, given an input vector \(x_i\), \(Q_i=W^Q x_i, K_i=W^K x_i, V_i=W^V x_i\), where \(W^Q,W^K,W^K\) are trainable weight matrices \cite{jalammargithub}. Two score functions were used, one from Bahdanau \cite{bahdanau2016neural}:
\begin{equation}
\mathrm{Bahdanau Score}= V^T \tanh{(K+Q)}
  \label{eq:bahdanau}
\end{equation}
and one from Luong \cite{LuongPM15}
\begin{equation}
\mathrm{Luong Score} = K^T Q
  \label{eq:luong}
\end{equation}
Following the work of \cite{scacnnchen2017}, we performed self-attention across channels. Given an image \(I  \in \mathbb{R}^{H \times L \times D}\), where \(H,L,D\) are the height, length and depth of the input images after a few layers of convolutions, for each \((h,w) \in \mathbb{R}^{H \times W}\), we performed feature-wise, or equivalently channel-wise attention, by performing the self-attention on the corresponding channel vector \(\in \mathbb{R}^D\), for both Luong and Bahdanau scoring.

  To perform self-attention across spatial dimensions, we used Transformers \cite{VaswaniSPUJGKP17allyouneed}. The Transformer consisted of alternating fully connected and multi-head attention (MHA) layers, the latter being multiple attention layers in parallel, whose outputs were concatenated to produce the output of the MHA layer. Given weight matrix \(W^O\):
  \begin{equation}
\mathrm{MultiHead} (Q,K,V) = \mathrm{concat}(h_0,h_1...h_i) W^O
  \label{eq:multihead}
\end{equation}
Where each \(h_j\) is the output of an Attention layer with its own set of weights:
\begin{equation}
h_j=Attention(QW^Q_j,KW^K_j,VW^V_j)
  \label{eq:more_attention}
\end{equation}
In this paper, the Attention function used Luong scoring, as is standard in the literature. \cite{visiontrans} introduced the Vision Transformer (ViT), which performs patch embedding by reshaping \(x \in \mathbb{R}^{H \times L \times D} \rightarrow x_p \in \mathbb{R}^{\frac{HL}{P^2} \times P^2 D}\), where  \(P \times P\) is the size of each patch in pixels. This patch embedded input is then passed to a transformer. 

\subsection{Residual Networks}
Increasingly, deeper and larger convolutional networks have been used for vision tasks \cite{Simonyan2015VeryDC,Krizhevsky2012ImageNetCW}. However, this led to the degradation problem, where accuracy for classification tasks would saturate after a certain level of depth. To solve this, \cite{he2015deep} proposed the residual connection, where the input of past layers would be added to the output of past layers. \cite{Philipp2018GradientsE} proved that ResNets were also capable of circumventing the problem of exploding gradients. \cite{Xie2017AggregatedRT} further improved on this by introducing the ResNeXt, which offered superior accuracy to ResNet models with the same number of parameters. The output of a ResNeXt layer \(y\) with input \(x\) is
\begin{equation}
y =x + \sum_{i}^C T_i(x)
  \label{eq:residual}
\end{equation}
Where each \(T_i\) is a transformation, \(C\), known as the cardinality, is the amount of transformations to be applied. This is equivalently implemented as the concatenation of the outputs of convolutional layers, an aggregation of residual layers, or a grouped convolution \cite{Krizhevsky2012ImageNetCW}.

\section{Visualization} \label{sec:appendixB} 

\begin{figure}[h]

\begin{subfigure}[]{0.45\linewidth}
\includegraphics[width=\linewidth]{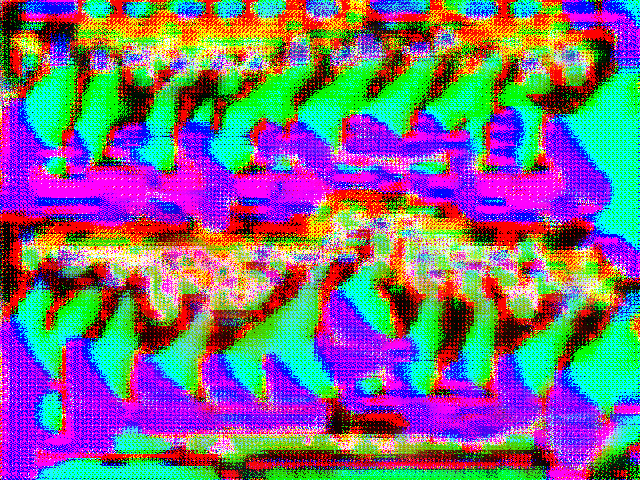} 
\label{fig:subim1}
\end{subfigure}
\begin{subfigure}[]{0.45\linewidth}
\includegraphics[width=\linewidth]{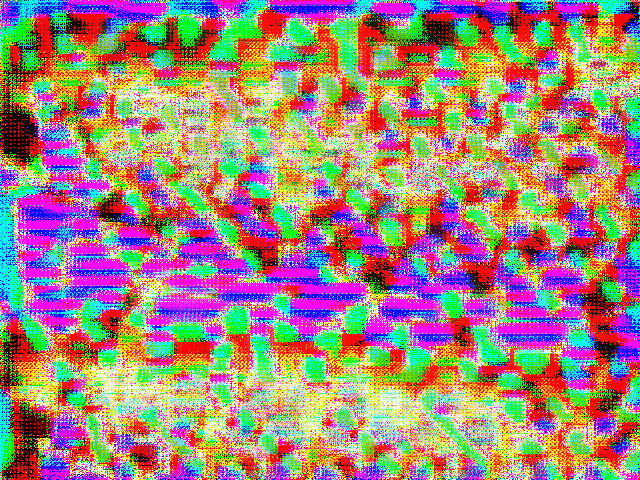}
\label{fig:subim2}
\end{subfigure}

\begin{subfigure}[]{0.45\linewidth}
\includegraphics[width=\linewidth]{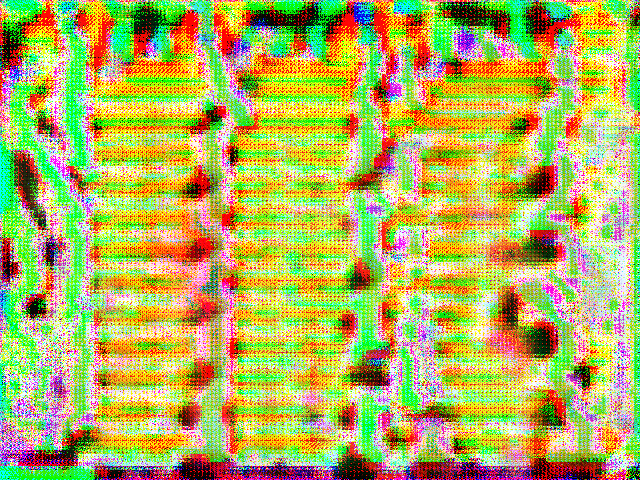} 
\label{fig:subim3}
\end{subfigure}
\begin{subfigure}[]{0.45\linewidth}
\includegraphics[width=\linewidth]{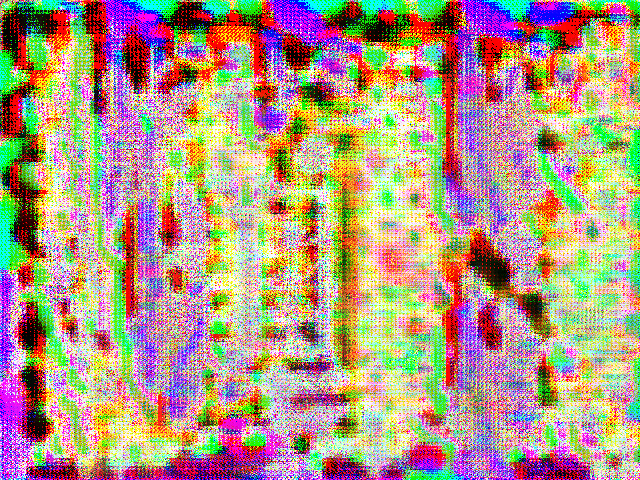}
\label{fig:subim4}
\end{subfigure}

\caption{Images generated for activation maximization of second to last layer of Model A-1 using gradient ascent}
\label{fig:resnext_layer_4}
\end{figure}

\begin{figure}[h]

\begin{subfigure}[]{0.45\linewidth}
\includegraphics[width=\linewidth]{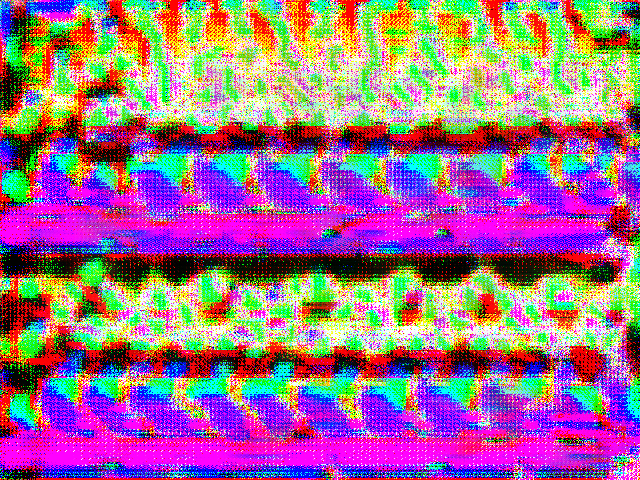} 

\end{subfigure}
\begin{subfigure}[]{0.45\linewidth}
\includegraphics[width=\linewidth]{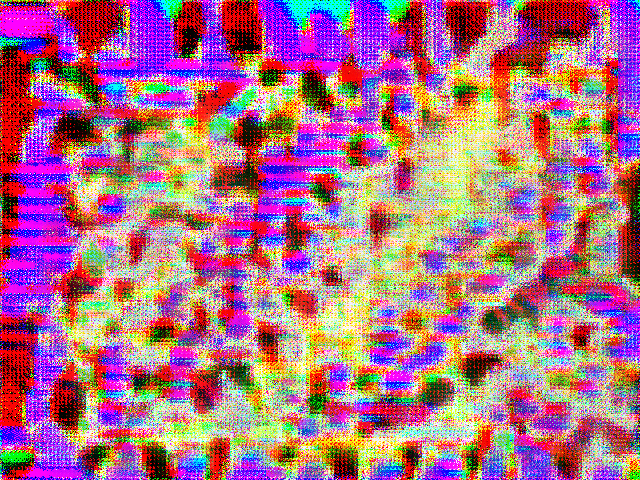}

\end{subfigure}

\begin{subfigure}[]{0.45\linewidth}
\includegraphics[width=\linewidth]{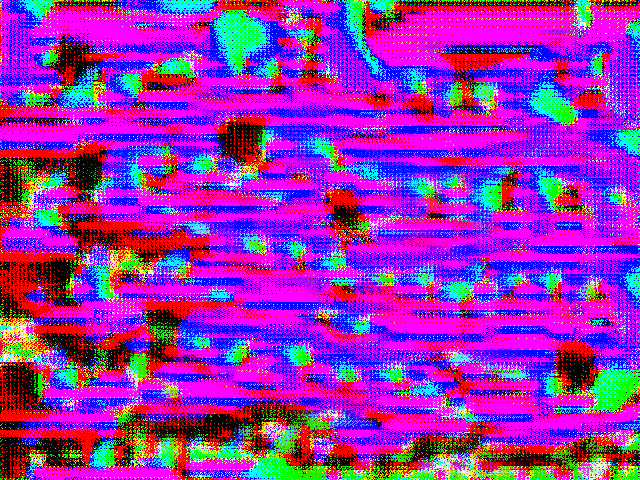} 

\end{subfigure}
\begin{subfigure}[]{0.45\linewidth}
\includegraphics[width=\linewidth]{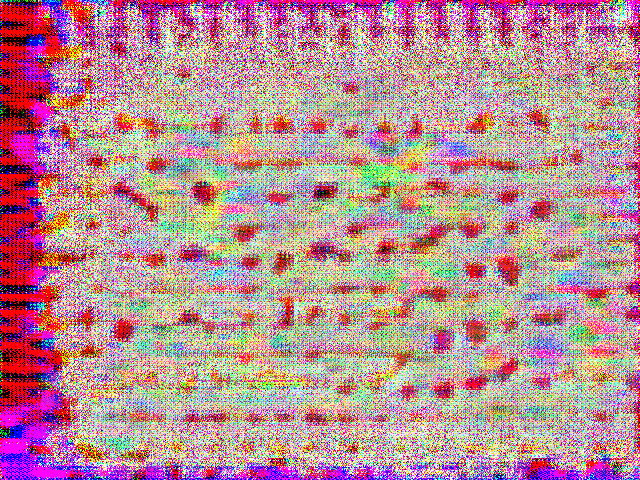}

\end{subfigure}

\caption{Images generated for activation maximization of last layer using gradient ascent}
\label{fig:resnext_layer_5}
\end{figure}

\end{document}